\documentclass[conference]{IEEEtran}
\ifCLASSINFOpdf
\else
\fi
\usepackage{amsmath, amsfonts, amssymb, graphicx, float}
\usepackage{amsbsy}
\usepackage{mathrsfs}
\usepackage{bm}
\usepackage{array,booktabs}
\usepackage{verbatim}
\usepackage{multirow}

\newcommand{\like}{\log P} 

\newcommand{\abs}[1]{\left| #1 \right| }

\newcommand{\dP}{\mathrm{d}p}

\newcommand{\Poi}{\text{Poi}}

\renewcommand{\mid}{\,|\,}

\begin{document}
%
\title{Bayesian Model Selection \\of Stochastic Block Models}

\author{\IEEEauthorblockN{Xiaoran Yan}
\IEEEauthorblockA{Indiana University Network Science Institute\\
Bloomington, Indiana 47408\\
Email: xiaoran.a.yan@gmail.com}
}


%


\maketitle

\begin{abstract}
A central problem in analyzing networks is partitioning them into \emph{modules} or \emph{communities}. One of the best tools for this is the \emph{stochastic block model}, which clusters vertices into blocks with statistically homogeneous pattern of links. Despite its flexibility and popularity, there has been a lack of principled statistical model selection criteria for the stochastic block model. Here we propose a Bayesian framework for choosing the number of blocks as well as comparing it to the more elaborate degree-corrected block models, ultimately leading to a universal model selection framework capable of comparing multiple modeling combinations. We will also investigate its connection to the minimum description length principle.
\end{abstract}


%
\IEEEpeerreviewmaketitle

\section{Introduction}
An important task in network analysis is community detection, or finding groups of similar vertices which can then be analyzed separately \cite{fortunato}. Community structures offer clues to the processes which generated the graph, on scales ranging from face-to-face social interaction \cite{zachary} through social-media communications \cite{adamic_pBlogs} to the organization of food webs \cite{FWpitfall}. However, previous work often defines a ``community'' as a group of vertices with high density of connections within the group and a low density of connections to the rest of the network.  While this type of \emph{assortative} community structure is generally the case in social networks, we are interested in a more general definition of \emph{functional} community---a group of vertices that connect to the rest of the network in similar ways.  A set of similar predators form a functional group in a food web, not because they eat each other, but because they feed on similar prey.  In English, nouns often follow adjectives, but seldom follow other nouns.

The \emph{stochastic block model} (SBM) is a popular network model for such functional communities \cite{blockmodel1}. It splits vertices into latent blocks, within which all vertices are stochastically equivalent in terms of how they connect to the rest of the network \cite{WASSERMAN:1987}. As a generative model, it has a well-defined likelihood function with consistent parameter estimates. It also provides a flexible base for more general latent state models. In particular, Karrer and Newman proposed a variant called the degree-corrected block models (DC-SBM) capable of capturing specific degree distributions and Yaojia et al. generalized it to directed graphs \cite{karrer,deggen}. There are also other variants capable of modeling overlapping, hierarchical and even ``meta-data'' dependent communities~\cite{airoldi,hierarchical,newman_structure_2015}.

Performance of different latent state models vary under different scenarios. For SBM and its generalizations, picking the right model (\emph{model selection}), and in particular picking the right number of blocks (\emph{order selection}) is crucial for successful network modeling. Numerous statistical model selection techniques have been developed for classic independent data. Unfortunately, it has been a common mistake to use these techniques in network models without rigorous examinations of their compatibility, ignoring the fact that some of the fundamental assumptions has been violated by moving into the domain of relational data \cite{neuroMS, pitfall, ecolPitfall}. As a result, some employed these information criteria directly with out knowing the consequences \cite{FWpitfall, socialPitfall}, while others remain skeptical and use it only when no alternatives is available \cite{airoldi, mixedMS}.

Our main contribution in this paper is to develop a Bayesian model selection framework for comparing multiple SBM variants with different number of blocks\footnote{We will focus on the model/order selection problems of the SBM and Degree-corrected block model. For mathematical convenience, we will define these models for undirected graphs. Directed versions require additional specifications but the general form remains the same \cite{deggen}. Readers should be able to generalize the result in this paper to these cases.}. In Section 2, we will first go over the Bayesian model selection approaches and the minimum description length (MDL) principle \cite{grunwald2007, peixoto2012}. In Section 3, we will propose Bayesian order selection for the SBM, leading to a Bayesian Information criterion (BIC) \cite{modelSA}. We will also establish the equivalence between BIC and MDL. In section 4, we will generalize these results to the DC-SBM, leading to a universal model selection framework capable of comparing multiple models that combines different model and order choices. We will compare its theoretic and empirical results to previous MDL based approaches \cite{peixoto2012,peixoto_hierarchical_2014} as well as previous work on 
frequentist model selection \cite{MS-DC}.

\section{Background and related work}
Model selection is about balancing the trade-off between model complexity and fit to the data. Models with more parameters have a natural advantage at fitting data. Simpler models have lower variability, and are less sensitive to noise in the data. A good model choice should avoid both over-fitting and under-fitting, and only include additional parameters when they do capture meaningful structures in the data \cite{modelSA, modelS}.

The frequentist approach to model selection cast the problem as a hypothesis testing. It focus on estimating the likelihood-ratio between a pair of candidate models under the null hypothesis. Such frequentist method has been used for both model selection and order selection \cite{MS-DC,wang_likelihood-based_2015}. In this paper, we shall follow the other major school of statistics and use Bayesian techniques for model selection.

\subsection{Bayesian model selection}
While point estimates like the maximum likelihood lead to over-fitting, Bayesian approaches take the whole posterior distribution into account, thus achieving the trade-off between model complexity and its fit to data~\cite{modelSA, modelS}. These posteriors distributions can be formulated using Bayes' rule,
\begin{align}
\label{eq:bayes-general}
P(M_i \mid G) =& \dfrac{P(M_i)}{P(G)} P(G \mid M_i) \nonumber\\
\propto& \iiint_0^1 P(G \mid M_i, \Pi_i) P(\Pi_i\mid M_i){\rm d} \Pi_i \;,
\end{align}
where we have assumed a uniform prior of models $P(M_i)$, and the total evidence of data $P(G)$ is constant for all models. 

The posterior $P(M_i \mid G)$ has an intuitive interpretation for model selection. It is proportional to $P(G \mid M_i)$, which is the integrated complete likelihood (as ICL in \cite{biernacki_assessing_2000}) $P(G \mid M_i, \Pi_i)$ over the prior of parameters $P(\Pi_i\mid M_i)$. To compare models, the standard approach is to divide one posterior with another in a likelihood-ratio style, leading to the Bayes factor \cite{kass_bayes_1995}. Unlike the frequentist tests, Bayes factor uses the ICL without any dependence on parameters. Therefore, it can be applied not only to nested model pairs, but two models of any form.

Ironically, instead of one preferred choice, a fully Bayesian approach would give a posterior distribution over all candidate models. The maximum a posteriori (MAP) method presented in this paper is technically a hybrid between Bayesian and frequentist ideas. For comparison with frequentist and MDL based methods, we will still call it Bayesian. Nonetheless, most results can be adapted for full Bayesian analysis.

\subsection{Bayesian information criterion}
While Bayesian model selection is compatible for models of any form, the posteriors can often be intractable. The exception is when the likelihood model is from a family with conjugate priors. The posteriors will then have closed form solutions. Fortunately for us, the block models fall under this category.

Bayes factor is also quite cumbersome when more than two candidate models are involved. Instead of the principled hypothesis testing framework,  Bayesian information criterion (BIC) \cite{schwarzBIC} gives standard values for each candidate model:
\begin{equation}
\label{eq:BIC2}
 \textit{BIC}(M_i) = -2\ln P(Y\mid M_i, \hat{\Pi}_i) + |\Pi_i|\ln n \, ,
\end{equation}
where $|\Pi_i|$ is the degree of freedom of the model $M_i$ with a parameter set $|\Pi_i|$, and $n$ is number of i.i.d. samples in the data. The above simple formulation consists of a maximized likelihood term and a penalty term for model complexity, intuitively corresponding to the trade-off we are looking for. As we will later show, it is in fact a large sample approximation to twice the logarithm of the ICL (Equation \eqref{eq:bayes-general}).

BIC has been applied to different clustering models of i.i.d. data \cite{fraley_how_1998,biernacki_assessing_2000,biernacki_exact_2010}. Handcock et al. proposed a variants of BIC for order selection of a latent space clustering model on networks \cite{socialPitfall}. Recently, C\^{o}me and Latouche derived a BIC from the ICL for the vanilla SBMs \cite{come_model_2013}. We will redo the derivation in our notions and generalize it to degree-corrected block models in this paper.

\subsection{The minimum description length principle}
By compressing data with different coding schemes, information theory has a long history dealing with the trade-off between complexity and fit. 
Searching for the model with best predictive performance is essentially finding a coding scheme that lead to the minimum description length (MDL)~\cite{grunwald2007, rosvall}. Under the MDL principle, the trade-off takes the from of balancing between the description length of the coding scheme and that of the message body given the code \cite{peixoto2012, peixoto_entropy}. 


MDL is closely related with Bayesian model selection, particularly the BIC formulation~\cite{grunwald2007}. In~\cite{peixoto_hierarchical_2014,peixoto_model_2014}, Peixoto demonstrated that for basic SBMs, the ICL or Bayesian posteriors are mathematically equivalent to MDL criteria under certain model constrains. This equivalence, as we will show in this paper, underlies a fundamental connection in the form of carefully designed Bayesian codes, which can be derived from the ICL~\eqref{eq:bayes-general} with intuitions.

\section{Bayesian order selection of the SBM}
In this section, we will formally introduce the SBM, derive the ICL for the SBM order selection problem, and design an intuitive Bayesian code to demonstrate the fundamental connection between Bayesian order selection and the MDL. Finally we will propose a Bayesian Information criterion (BIC) \cite{modelSA} for order selection of the SBM. 

\subsection{The stochastic block model}

We represent our network as a simple undirected graph $G=(V,E)$,  without self-loops. $G$ has $n$ vertices in the set $V$, $m$ edges in the set $E$, and they can be specified by an adjacency matrix $A$ where each entry $A_{uv} = 0$ or $1$ indicates if there is an edge in-between. We assume that there are $k$ blocks of vertices (choosing $k$ is the order selection problem), so that each vertex $u$ has a block label $g(u) \in \{1,\ldots,k\}$. Here $n_s = \abs{ \{u \in V : g(u) = s \} }$ is the number of vertices in block $s$, and $m_{st} = \abs{ \{u<v, (u,v) \in E : g(u) = s, g(v) = t \} }$ is the number of edges connecting between block $s$ and block $t$.

We assume that $G$ is generated by a SBM, or a ``vanilla SBM'' as we will call it throughout this paper for distinction. For each pair of vertices $u,v$, there is an edge between $u$ and $v$ with the probability $p_{g(u),g(v)}$ specified by the $k \times k$ block affinity matrix $p$. Each vertex label $g(u)$ is first independently generated according to the prior probability $q_{g(u)}$ with $\sum_{s=0}^k q_s = 1$. Given a block assignment, i.e., a function $g:V \to \{1,\ldots,k\}$ assigning a label to each vertex, the probability of generating a given graph $G$ in this model is
\begin{align}
\label{eq:like-tp}
&P (G,g \mid q,p)\nonumber\\
=& \prod_{s=1}^k q_i^{n_s} \prod_{s\le t}^k p_{st}^{m_{st}} (1-p_{st})^{n_s n_t - m_{st}}\, ,
\end{align}
This likelihood factors into terms for vertices and edges, conditioned on their parameters $q,p$ respectively.

Take the log of Equation \eqref{eq:like-tp}, we have the log-likelihood 
\begin{align}
&\like(G,g \mid q,p) = \sum_{s=1}^k{n_s\log{q_s}} \nonumber\\
&+ \sum_{s\le t}^k \left( m_{st}\log p_{st} + (n_s n_t - m_{st})\log(1-p_{st}) \right) \,.
\label{eq:ll-tp}
\end{align}

\subsection{Bayesian posterior of the SBM}
A key design choice of MAP model selection is picking which parameters to integrate. By being partially frequentist, we have the freedom to decide how "Bayesian" we wanted to be. The decision can also be understood as balancing between the bias in the learning task and the variance in application domains. For example, if the learning task is to find the latent state model with the most likely parametrization, regardless of specific latent block assignments, the integral over parameters is not necessary, as in the paper~\cite{MS-DC}.

Alternatively, if the learning task is to find the model with the most likely latent state, we do not need the sum over the latent state $g$, and benefit from the smaller variance because of the bias we are willing to assume. However, if we plan to apply the learned latent state model to similar networks with different parameterizations, the integral over parameters remains essential. This corresponds to the idea of \emph{Universal Coding} in information coding theories \cite{grunwald2007}, where a code has to achieve optimal average compression for all data generated by the same code scheme without knowing its parametrization a priori. This is the approach Handcock et al. adopted for order selection of their latent space model~\cite{socialPitfall}. The same method was used in the active learning algorithm for vanilla SBMs~\cite{moore_active}. In this paper, we will follow this approach for its connection with MDL methods. We will use a Monte Carlo sampling method to find the most likely latent state $\hat{g}$.

According to equation \eqref{eq:bayes-general}, we have the posterior of a SBM $M_i$ with the parameters $\{p, q\}$:
\begin{align}
\label{eq:fullBayes}
 P(M_i \,|\, G) =& \dfrac{ P(M_i)}{ P(G)}  P(G \,|\, M_i) \nonumber\\
	  \propto& \sum_g \iiint_0^1 {\rm d}\{p_{st}\} {\rm d}\{ q_i\}\, P(G,g\,|\,p, q)\;,
\end{align} 
where we assume the prior of models $ P(M_i)$ is uniform, and the total evidence of data $ P(G)$ is constant. Here the ICL $ P(G \,|\, M_i)$ is integrated over both $p$ and $ q$ entries, as well as summed over all latent states $g$. Since we are interested in the most likely latent state $\hat{g}$, we can forgo the sum,
\begin{align}
  P(G ,g\mid M_i) = \iiint_0^1 {\rm d}\{p_{st}\} {\rm d}\{q_s\}\, P(G , g\mid q,p)\;.
\end{align}
with $P(G ,\hat{g}\mid M_i) = \max_{g} P(G ,g\mid M_i)$.

If we assume that the $p_{st}$ and $q_s$ entries are independent, conditioned on the constrain $\sum_sq_s=1$, and they follows their respective conjugate Dirichlet and Beta priors ($\delta$ and $\{\alpha, \beta\}$ respectively), we have
\begin{align}
\label{eq:Bayes}
& P(M_i \mid G) \propto  \iiint_0^1 {\rm d}\{p_{st}\} {\rm d}\{q_s\}\, P(G ,g\mid q,p) \nonumber \\
=& \left( \int_\triangle {\rm d} q \mathrm{Dirichlet}(\vec{q}|\vec{\delta}) \prod_{s=1}^k q_s^{n_s}  \right)\nonumber \\
 & \left( \prod_{s\le t}^k \int_0^1 \dP_{st} \,\mathrm{Beta}(p_{st}|\alpha, \beta) \,p_{st}^{m_{st}} (1-p_{st})^{n_s n_t - m_{st}}\right) \nonumber \\
=& \left( \dfrac{ \Gamma (\sum_{s=1}^k \delta_s)}{\prod_{s=1}^k \Gamma(\delta_s)} \dfrac{\prod_{s=1}^k \Gamma(n_s+\delta_s)}{\Gamma(\sum_{s=1}^k (n_s+\delta_s))}  \right)\nonumber \\
 & \left(\prod_{s\le t}^k \dfrac{ \Gamma (\alpha+\beta)}{\Gamma(\alpha) \,\Gamma(\beta)} \dfrac{\Gamma(m_{st}+\alpha) \,\Gamma(n_s n_t-m_{st}+\beta)}{ \Gamma (n_s n_t+\alpha+\beta)}\right)\nonumber\\
=&  P(V ,g\mid M_i) \times  P(E ,g\mid M_i) \;,
\end{align}
where we have assumed the same beta prior $\{\alpha, \beta\}$ for all $p_{st}$ entries, and applied the Euler integral of the first kind, and its multinomial generalization on the simplex $\sum_sq_s=1$. 

Equation~\eqref{eq:Bayes} shows, the ICL factors into terms for vertices and edges. It holds for any latent state $g$, as long as $n_s$ and $m_{st}$ terms are consistent with the given $g$. For empirical verifications, we will use a Monte Carlo sampling method to find the most likely latent state $\hat{g}$. 

To get an idea of the posterior distribution $P(M_i \mid G)$, we assume that the data $G$ follows a uniform prior over random graphs generated by a SBM with 5 prescribed blocks. For simplicity, we have also plugged in the uniform priors (i.e., $\delta_{\forall s}=1,\alpha=\beta=1$) for the parameters, as it is done in~\cite{moore_active},
\begin{align}
\label{eq:Bayes_uniform}
& P(M_i \mid G,g)\propto P(G ,g\mid M_i)\nonumber \\
=& \left((k-1)! \dfrac{\prod_{s=1}^k n_s!}{(n+k-1)!} \right)\left( \prod_{s\le t}^k \frac{m_{st}! (n_s n_t -m_{st})!}{(n_s n_t + 1)!} \right) \;.
\end{align} 

\begin{figure}
\begin{center}
\includegraphics[width=.4\textwidth]{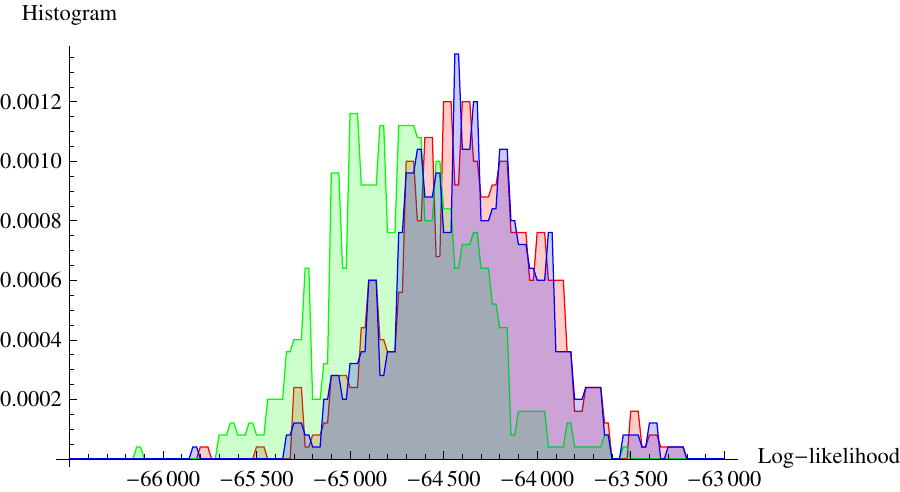}
\includegraphics[width=.4\textwidth]{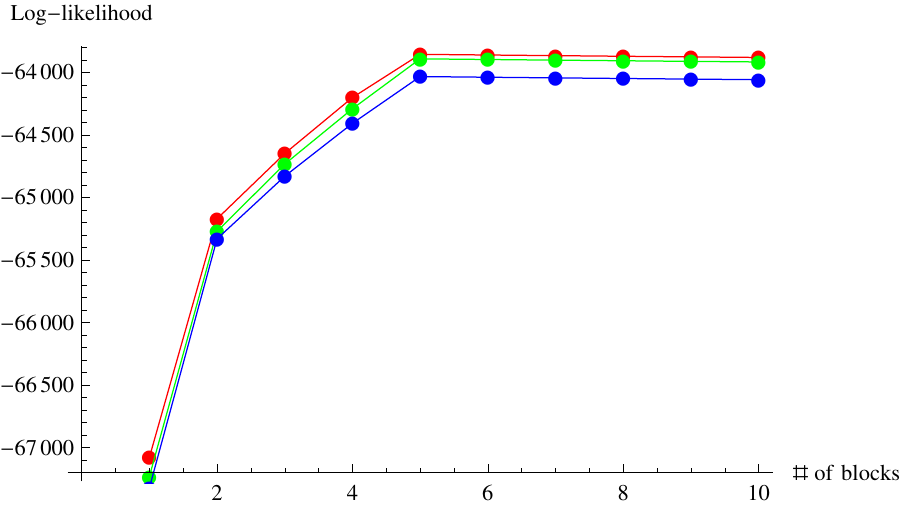}
\end{center}
\caption{Top: the histogram of log-integrated complete likelihoods (log-ICL) of the Bayesian model (Equation \eqref{eq:Bayes_uniform}). The distributions are gathered from randomly generated SBMs with 1000 vertices and 5 prescribed blocks. SBMs with different number of blocks $k$ are fitted to the data. Specifically, the green distribution is from a SBM with $k=4$, the red with $k=5$ and the blue with $k=9$. The experiment is done using Monte Carlo sampling, with 500 total samples of $G$. Bottom: the change of log-ICL for sample graphs as the number of blocks grows. Three graphs are randomly generated according to the same prescribed blocks. All of them have highest log-likelihood at the correct number of blocks $k=5$.}
\label{fig:bayesDist}
\end{figure}

The distributions of the posterior with different number of blocks are shown in Figure \ref{fig:bayesDist} (top). While the SBM with correct number of blocks (red) does has slightly higher likelihood in average, it overlaps quite heavily with SBMs with fewer blocks (green) or more blocks (blue).  Further investigation reveals that most of the variance came from the randomness in the generated data. Once we fix the input graph for all the candidate models, the SBM with correct number of blocks always has a higher likelihood than the others, as illustrated in Figure.~\ref{fig:bayesDist} (bottom).

\subsection{Bayesian code for the SBM}
\label{sec:codeSBM}
Now we will design a Bayesian code based on the ICL. According to Gr\"{u}nwald \cite{grunwald2007}, the equivalence between Bayesian and the MDL principle for model selection holds in general. In our case, if we choose the Jeffreys priors for $p_{st}$ and $q_s$ entries (i.e., $\alpha=\beta=\delta_{\forall s}=1/2$), the coding length according to the Bayesian model is asymptotically the same as the optimal minimax coding. Gr\"{u}nwald also pointed out in \cite{grunwald2007}, while the Jeffreys priors lead to the optimal universal coding, other priors and their corresponding non-optimal coding still lead to description length of the same asymptotic order, if the prior is dominated by the evidence.

This justifies the uniform prior assumptions we used in Equation \eqref{eq:Bayes_uniform}, and we can rewrite it as
\begin{align}
\label{eq:combinatorial}
& P(G ,g\mid M_i) =  P(V ,g\mid M_i) \times  P(E ,g\mid M_i) \nonumber \\
=& \left(\dfrac{1}{{n+k-1\choose k-1}}\dfrac{1}{{n\choose (n_1,n_2,...,n_k)}} \right)\left( \prod_{s\le t}^k \dfrac{1}{{n_s n_t\choose m_{st}}(n_s n_t + 1)} \right) \;.
\end{align} 

The dominating terms in Equation \eqref{eq:combinatorial} lead to a Bayesian universal code for a graph $G$ consists of the following parts:
\begin{enumerate}
 \item number of blocks $k$ ($\log k$ bits, implicit)
 \item code for the partition of $n$ into $k$ $n_s$ terms ($\log{n+k-1\choose k-1}$ bits, ), which will specify a block size sequence (ordered in terms of blocks)
 \item code for assigning each vertex to blocks according to the $n_s$ terms ($\log{n\choose (n_1,n_2,...,n_k)}$ bits)
 \item for each pair of blocks ${s,t}$, the number of undirected edges $m_{st}$ going between  block $s$ and block $t$ ($\log m_{st} < \log (n_s n_t + 1)$ bits)
 \item for each pair of blocks ${s,t}$, code for the edge allocations given $m_{st}$ ($\log{n_s n_t\choose m_{st}}$ bits, uniformly random allocations)
\end{enumerate}

According to \cite{grunwald2007}, a mapping exists between probability distributions and prefix codes. In the Bayesian code, the distribution of possible realizations in part $i$ ($i >1$) conditioned on all previous code parts are all uniform, the optimal code length for part $i$ thus can be quantified by the negative logarithm of the corresponding combinatorial terms in Equation \eqref{eq:combinatorial}. 

The aforementioned Bayesian code is identical to the description length for the single level model discussed in \cite{peixoto_hierarchical_2014}. In fact, Peixoto mathematically arrived at the same equivalence in the appendix of \cite{peixoto_hierarchical_2014}. Earlier formulations of MDLs, however, are usually defined in terms of the entropy minimizers of the likelihood functions, which is equivalent to the maximum likelihood formulation in the BIC (Equation \eqref{eq:BIC2}).

\subsection{BIC for order selection of SBMs}
\label{sec:BIC-SBM}
The key to transform Equation~\eqref{eq:Bayes} to the BIC formulation (Equation~\eqref{eq:BIC2}) is the Laplace's approximation with uniform priors. Or equivalently, by using Stirling's formula on the factorials in Equation \eqref{eq:combinatorial},
\begin{align}
\label{eq:Stirling}
& P(G,g \mid M_i) =  P(V ,g\mid M_i) \times  P(E ,g\mid M_i) \nonumber \\
	\approx & \dfrac{\prod_{s=1}^k\sqrt{2\pi n_s}}{{(n+k-1)\choose n}\sqrt{2\pi n}} \prod_u \dfrac{n_{g(u)}}{n} 
		  \prod_{s\le t}^k \dfrac{2\pi \sqrt{m_{st}(n_s n_t -m_{st})}}{\sqrt{2\pi (n_s n_t)}(n_s n_t+1)}\nonumber\\
		& \prod_{u<v, (u,v) \in E}\dfrac{m_{g(u)g(v)}}{n_{g(u)} n_{g(v)}} \prod_{u<v,(u,v)\notin E}(1-\dfrac{m_{g(u)g(v)}}{n_{g(u)} n_{g(v)}}) \nonumber\\
	\approx &  P(V ,g\mid \hat{q})\dfrac{\prod_{s=1}^k\sqrt{2\pi n_s}}{{(n+k-1)\choose n}\sqrt{2\pi n}} \nonumber\\
		&\times P(E ,g\mid \hat{p})\prod_{s\le t}^k \dfrac{\sqrt{2\pi}}{\sqrt{\dfrac{n_s^3n_t^3}{m_{st}(n_sn_t-m_{st})}}} \; ,
\end{align}
where we plugged in the MLEs $\hat{q_s} = \dfrac{n_{s}}{n} $ and $\hat{p_{st}}=\frac{m_{st}}{n_s n_t}$. 

If we take the negative log of Equation \eqref{eq:Stirling}. The factor associated with $E$ becomes:
\begin{align}
& -\ln  P(E ,g\mid M_i) \nonumber \\
\approx& -\ln   P(E ,g\mid \hat{p}) - \sum_{s\le t}^k \frac{1}{2}\ln \dfrac{2\pi m_{st}(n_sn_t-m_{st})}{n_s^3n_t^3}\nonumber \\
\approx& -\ln   P(E ,g\mid \hat{p}) + \frac{k^2}{2}\ln \dfrac{n^6}{4\pi |E|(n^2-|E|)} -C\;,
\end{align}
where we made a mean-field assumption about both $n_s$ and $m_{st}$ under constant number of blocks $k$. If the edge density of graph scales as $|E| = \rho n^2$, with $\rho$ being a constant such that $0\leq \rho \ll 1$, we have
\begin{align*}
-\ln  P(E ,g\mid M_i) \approx& -\ln  P(E ,g\mid \hat{p}) + \frac{k^2}{2}\ln \dfrac{n^6}{4\pi \rho n^4(1-\rho)} \\
			  \approx& -\ln P(E ,g\mid \hat{p}) + \frac{k^2}{2}\ln \Theta(n^2) \;.
\end{align*}

Putting it together with the term associated with $V$, in which we again assumed mean-field $n_s$ terms, we get
\begin{align}
\label{eq:sumBayes}
&-\ln  P(G ,g\mid M_i) = -\ln P(V ,g\mid M_i) - \ln P(E ,g\mid M_i) \nonumber \\
\approx & -\ln P(V ,g\mid \hat{q}) +\Theta(k \ln n) -\ln P(E ,g\mid \hat{p}) +\frac{k^2}{2}\ln \Theta(n^2) \nonumber\\
      = & -\ln P(G ,g\mid \hat{p},\hat{q}) + \frac{k^2}{2}\ln \Theta(n^2) \;.
\end{align}

Multiply by $2$, we have the BIC for order selection in SBMs:
\begin{align}
\label{eq:BIC_dense}
\textit{BIC}_{SBM}(M_i) = - 2\ln P(G,g|M_i, \hat{\Pi}_i) + k^2\ln \Theta(n^2)\;,
\end{align}
with $k^2$ specifying the number of parameters in the block affinity matrix $p$ and $n^2$ represent the sample size as pairwise edge/non-edge interactions.

\eqref{eq:BIC_dense} is simply the direct application of BIC to the SBM as it is defined in Equation \eqref{eq:like-tp}. In \cite{socialPitfall}, Handcock et al. arrived at the same equation without showing derivations. They also suggested using $|E|$ instead of $n^2$ as the sample size measure, on the basis of arguments that apply to a very different model.
However, the above derivation is no longer correct when the edge density scales differently. If the graph is sparser with $|E| = \rho n$, we have
\begin{align}
\label{eq:sumBayesSparse}
-\ln  P(E ,g\mid M_i) \approx& -\ln  P(E ,g\mid \hat{p}) + \frac{k^2}{2}\ln \dfrac{n^6}{4\pi \rho n^2(n-\rho)} \nonumber\\
			  \approx& -\ln P(E ,g\mid \hat{p}) + \frac{k^2}{2}\ln \Theta(n^3) \;.
\end{align}
The sparse BIC for order selection is then:
\begin{align}
\label{eq:BIC_sparse}
\textit{BIC}^{\ *}_{SBM}(M_i) = - 2\ln P(G,g|M_i, \hat{\Pi}_i) + k^2\ln \Theta(n^3)\;,
\end{align}
where the penalty term becomes even greater, favoring simpler models to compensate for sparser data samples (edges).

\section{Bayesian model selection of the DC-SBM}
In this section, we will generalize the results to the degree-corrected block model, ultimately leading to a universal model selection framework capable of comparing multiple models with different number of blocks.

\subsection{The Degree-corrected block model}
\label{sec:P-DCSBM}
The vanilla SBM assumes that each entry $A_{uv}$ is $0$ or $1$. Another restriction of the vanilla SBM is that all the vertices in the same block have the same expected degree, following a Binomial distribution with a narrow peak. As a consequence, it ``resists'' putting vertices with very different degrees in the same block, leading to problems in real networks when the degree distribution is heavy tailed. 

The DC-SBM addresses these problem by allowing degree heterogeneity within blocks. Each vertex gets an additional parameter $\theta_u$, which scales the expected number of edges connecting to it \cite{karrer}. DC-SBM also generalizes the edge generating processs to Poisson, thus allowing multi-edges between vertices. According to the block assignment $g$, the means of these Poisson draws depends on the $k \times k$ block affinity matrix $\omega$, which replaces the $p$ matrix in the vanilla SBM. The edge generating likelihood is now
\[
A_{uv}|g \sim \Poi(\theta_u \theta_v \omega_{g_ug_v}) \,.
\]

The parameter $\theta_u$ gives us control over the expected degree of each vertex, which for instance, could be a measure of popularity in social networks. The likelihood stays the same if we scale $\theta_u$ for all vertices in block $s$, provided we also divide $\omega_{st}$ for all $s$ by the same factor. Thus identification demands an additional constraint. Here we use a convenient one that forces $\theta_u$ to sum to the total number of vertices in each block: $\sum_{u:g_u=s} \theta_u = n_s$. The ICL of DC-SBM is then
\begin{align}
  \label{eq:lh_DC}
& P(G,g\mid\theta,\omega,q) \nonumber\\
=& \prod_u{q_{g_u}}\prod_{u<v}{\frac{\left(\theta_u\,\theta_v\,\omega_{g_ug_v}\right)^{A_{uv}}}{A_{uv}!}\exp(-\theta_u\,\theta_v\,\omega_{g_ug_v})} \nonumber\\
			  =& \prod_u{\theta_u^{d_u} \prod_{s=1}^k{q_s^{n_s}} \prod_{s\le t}^k \omega_{st}^{m_{st}}\exp(- n_s n_t \omega_{st})}\prod_{u<v} \frac{1}{A_{uv}!}\\
			  =&  P(\Theta,g\mid \theta) \times P(V ,g\mid q) \times P(E ,g\mid p) \nonumber\, ,
\end{align}
where $d_u$ is the degree of vertex $u$, and $P(\Theta,g\mid \theta)$ is the only factor containing the $\theta$ parameters. 

DC-SBM can be simplified when modeling simple graphs. The last term becomes $1$, and if we take the logarithm,
\begin{align*}
& \like(G,g\mid\theta,\omega,q) \nonumber\\
=& \sum_u{d_u\log{\theta_u}} + \sum_{s=1}^k{n_s\log{q_s}} + \sum_{s\le t}^k(m_{st}\log{\omega_{st}}-n_s n_t \omega_{st}) \,. 
\end{align*}

Compare it with Equation \eqref{eq:ll-tp}, if the graph is not very dense such that $m_{st}\ll n_sn_t$,
\begin{align*}
&\like(G,g \mid q,p) = \like(G,g\mid 1, \omega,q) \\
=& \sum_{s=1}^k{n_s\log{q_s}} + \sum_{s\le t}^k ({m_{st}\log{p_{st}}-(n_s n_t -m_{st})\log(1-p_{st})} )\,.
\end{align*}

In other words, when $\omega_{st}\approx p_{st}$ and both approach $0$, multi-edges are so rare that the DC-SBM becomes the vanilla SBM by setting $\theta_u=1$ for all $u$. This nested model relationship is consistent with their Poisson counterparts in \cite{MS-DC}. For mathematical convenience, we shall automatically make these approximations in the following sections\footnote{Notice that it is different from the notion of sparsity in Equation \eqref{eq:BIC_sparse}. Even as $\omega_{st}\approx p_{st} \rightarrow 0$, we can still have quadratic scaling of edge densities.}. 

\subsection{Bayesian posterior of the DC-SBM}
We will now generalize the Bayesian framework to the DC-SBM (Equation \eqref{eq:lh_DC}). By Bayes' theorem, we have the posterior of a DC-SBM $M_i$ with the parameters $\{\theta,\omega,q\}$:
\begin{align}
 &P_{DC}(M_i \mid G) = \dfrac{ P(M_i)}{ P(G)}  P_{DC}(G \mid M_i) \nonumber\\
	     \propto& \iiint_0^1 {\rm d}\{\theta_u\}{\rm d}\{\omega_{st}\} {\rm d}\{q_s\}\, P(G ,g\mid \theta,\omega,q)\;,
\end{align} 
where we again assumed a uniform prior over $P(M_i)$, and a constant $ P(G)$.

If $\theta_u$, $w_{st}$ and $q_s$ entries are independent, with the constrains $\sum_{u:g_u=s} \theta_u = n_s$ and $\sum_iq_s=1$, we have the posterior:
\small
\begin{align}
\label{eq:Bayes_DC}
&P_{DC}(M_i \mid G,g) \propto  P_{DC}(G ,g\mid M_i)\nonumber\\
			=& \iiint_0^1 {\rm d}\{\theta_u\} {\rm d}\{\omega_{st}\} {\rm d}\{q_s\}\, P(G ,g\mid \theta,\omega,q) \nonumber \\
			=& \left( \int_\triangle {\rm d}\{\theta_u\} \prod_u \theta_u^{d_u}  \right) \iiint_0^1 {\rm d}\{\omega_{st}\} {\rm d}\{q_s\}\, P(G ,g\mid \omega,q) \nonumber \\
		  \approx& P(\Theta,g\mid M_i)\times P(G ,g\mid M_i)\;.
\end{align}
\normalsize
The integrated DC-SBM has one additional factor, forming a pair of nested models with the integrated vanilla SBM.

To prepare $P(\Theta,g\mid M_i)$ for Bayesian treatments, we first change the variables $\theta_u=n_{g(u)}\eta_u$ in the first integral, making the integrand a proper multinomial distribution. Now if the new parameters $\eta_u$ follow their Dirichlet conjugate priors,
\begin{align}
\label{eq:theta-eta}
&P(\Theta ,g\mid M_i)\approx \prod_{s=1}^k\left(\int_\triangle{\rm d}\{\eta_u\} \prod_{g(u)=s}\eta_u^{d_u}\right)\times\prod_u n_{g(u)}^{d_u+1}\nonumber \\
		     =& \prod_{s=1}^k\left(\int_\triangle {\rm d}\eta \mathrm{Dirichlet}(\vec{\eta_s}|\vec{\gamma_s})\prod_{g(u)=s}\eta_u^{d_u} \right) \times\prod_u n_{g(u)}^{d_u+1}\nonumber\\
		     =& \prod_{s=1}^k\left( (n_s-1)! \dfrac{\prod_{g(u)=s} d_u!}{(D_s+n_s-1)!} \right) \times\prod_u n_{g(u)}^{d_u+1}\;,
\end{align}
where we applied the multinomial Euler integral on the simplex $\sum_{u:g_u=s} \eta_u = 1$. At the last line of the derivation, we again assume the priors are uniform (i.e., $\gamma_{\forall u}=1$).

Following the derivation of BIC for the vanilla SBM, we apply the Stirling’s formula to the factorials in \eqref{eq:theta-eta},
\begin{align}
\label{eq:BIC_theta}
&P(\Theta ,g\mid M_i)\nonumber\\
	      \approx& P(\Theta ,g\mid \hat{\eta}) \prod_{s=1}^k\left(\dfrac{\prod_{g(u)=s} \sqrt{2\pi d_u}}{{(D_s+n_s-1)\choose D_s}\sqrt{2\pi D_s}} \right) \times\prod_{s=1}^k n_{s}^{D_s+n_s}\nonumber\\
		    =& P(\Theta ,g\mid \hat{\theta}) \prod_{s=1}^k\left(\dfrac{\prod_{g(u)=s} \sqrt{2\pi d_u}}{{(D_s+n_s-1)\choose D_s}\sqrt{2\pi D_s}} \right) \times\prod_{s=1}^k n_{s}^{n_s}\;,
\end{align} 
where $D_s = \sum_{g(u)=s} d_u$ is the total degree of vertices in block $s$. We have also plugged in the MLEs $\hat{\eta}_u=\frac{\hat{\theta}_u}{n_{g{u}}}=\frac{d_u}{D_{g(u)}}$.

Putting back the factors from the vanilla SBM (Equation \eqref{eq:sumBayes}), and take the logarithm of it, we have the log-ICL:
\begin{align}
\label{eq:BIC_DC_derive}
       & \ln P_{DC}(G ,g\mid M_i) = \ln P(\Theta,g\mid M_i)+\ln P(G ,g\mid M_i)\nonumber\\
      \approx& \ln P(G,g\mid\hat{\theta},\hat{q},\hat{p}) -\frac{k^2}{2}\ln \Theta(n^*)\nonumber\\
      &+ \Theta(\frac{n}{2}\ln \frac{n}{k})-\Theta(n\ln \frac{2|E|}{n})\,,
\end{align}
with again mean-field assumptions about $n_s$ and $D_s$ terms. Notice that $\Theta(n^*)$ is a general form for the correct sample size for graphs with different edge density scalings, corresponding to both Equation \eqref{eq:sumBayes} and \eqref{eq:sumBayesSparse}. 

\begin{figure}
\begin{center}
\includegraphics[width=.4\textwidth]{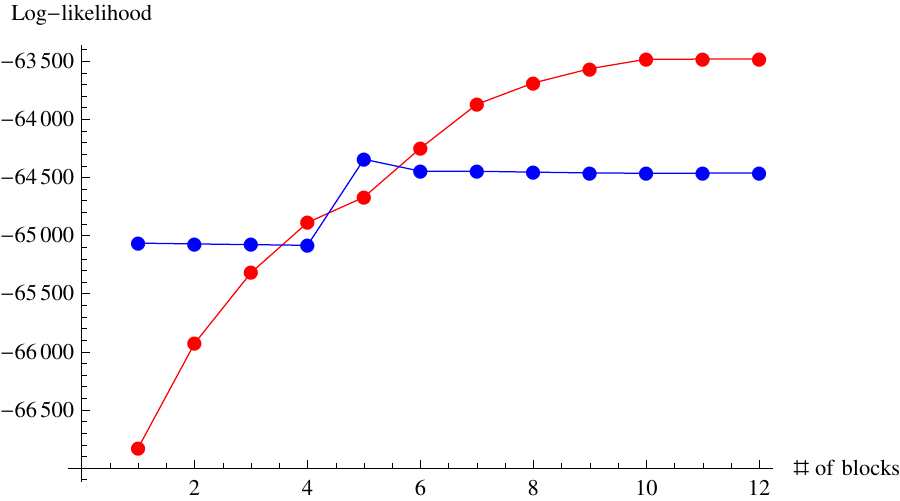}
\end{center}
\caption{The change of log-likelihood for given graphs as the number of blocks grows. The graph is a randomly generated DC-SBM with 1000 vertices and 5 prescribed blocks. Within each block, the degrees follow a bimodal distribution comprised of two Poisson distributions with their means 3 times apart. Both the vanilla SBM (red) and DC-SBM (blue) with different number of blocks $k$ are fitted to the data. The experiment is done using a Monte Carlo sampling method. The log-likelihood values shown here has been normalized for both models so that they are comparable across models. (The normalization method will be formally introduced in the next section.}
\label{fig:dcDistVdc}
\end{figure}

The blue curve in Figure \ref{fig:dcDistVdc} shows the empirical results. Here data $G$ is generated by a DC-SBM with $n=1000$ vertices from 5 prescribed blocks and the same expected number of total edges. Degrees within each block now follows a bimodal distribution. This degree heterogeneity forces the vanilla SBM to split vertices into separate high degree and low degree blocks, while the DC-SBM can comfortably mix them together in the same block. As a consequence, posterior of the DC-SBM achieves much higher log-ICL with fewer blocks. It also correctly captures the correct number of blocks at $k=5$, unlike the monotonic increasing log-ICL of the vanilla SBM.

\subsection{Towards a universal model selection framework}
In theory, a Bayesian approach based on the full ICL of Equation~\eqref{eq:fullBayes} can be used for comparing multiple SBM variants with different number of blocks together. However, our partially Frequentist approaches come with additional complications. In the previous example (Figure. \ref{fig:dcDistVdc}), we took an extra step to normalize the log-ICL for the vanilla SBM and DC-SBM so that they are comparable. The normalization step is required because our MAP approach leaves some parameters out of integration or summation. As a result, the ICL is still dependent of these parameters, eventually leading to divergence in maximum likelihoods for different models. In our case, these are the block assignment variables $g$, which can change drastically from the vanilla SBM to the DC-SBM given the same graph. 

To remedy the situation, we propose a normalization method that use a hybrid frequentist-Bayesian method by calculating the expected difference between Equation \eqref{eq:sumBayes}/\eqref{eq:sumBayesSparse} and \eqref{eq:BIC_DC_derive}:
\begin{align*}
	&\ln P_{DC}(G ,\hat{g}\mid M_i)-\ln P_{SBM}(G ,\hat{g}'\mid M_i') \\
 \approx& \ln P(G,\hat{g}\mid\hat{\theta},\hat{q},\hat{\omega}) - \ln P(G,\hat{g}'\mid 1,\hat{q}',\hat{p})\\
        &+\Theta(\frac{n}{2}\ln \frac{n}{k})-\Theta(n\ln \frac{2|E|}{n})\;,
\end{align*}
where $\hat{g}$ and $\hat{g}'$ are the most likely block assignments for the DC-SBM and vanilla SBM respectively.

Equation~\eqref{eq:Bayes_DC} shows that the vanilla SBM and the DC-SBM still forms a pair of nested models after the partial integrations. Therefore the analysis in \cite{MS-DC} holds. If we assume the underlying data is generated by the simpler vanilla SBM, we have both models converge to the same values for shared parameters, i.e. $\hat{g} = \hat{g}', \hat{q} = \hat{q}'$ and $\hat{\omega} = \hat{p}$. In \cite{MS-DC}, the authors came into the conclusion that the difference between the maximum likelihood under the null model roughly follows a $\chi^2$ distribution with a degree of freedom $n-k$, but corrections are needed when the graph is sparse.

To verify that we have the same maximum log-likelihood ratio, we can rewrite \eqref{eq:sumBayes} and \eqref{eq:BIC_DC_derive} as:
\begin{align*}
	   \ln P(G,g\mid 1,\hat{q},\hat{p}) \approx& \ln P_{SBM}(G ,g\mid M_i) + \frac{k^2}{2}\ln \Theta(n^*)\, ,\\
 \ln P(G,g\mid\hat{\theta},\hat{q},\hat{p}) \approx& \ln P_{DC}(G ,g\mid M_i) + \frac{k^2}{2}\ln \Theta(n^*)\\
					      &- \Theta(\frac{n}{2}\ln \frac{n}{k})+\Theta(n\ln \frac{2|E|}{n})\, .
\end{align*}

Therefore, we have the log-likelihood ratio,
\begin{align*}
 \Lambda_{DC}(G,g) =& \ln P(G,g\mid\hat{\theta},\hat{q},\hat{p}) - \ln P(G,g\mid 1,\hat{q},\hat{p})\\
	     \approx& \ln P(\Theta,g\mid M_i) - \Theta(\frac{n}{2}\ln \frac{n}{k})+\Theta(n\ln \frac{2|E|}{n})\\
		   =& \ln P(\Theta,g\mid \hat{\theta})\, ,
\end{align*}
which is exactly the same as the log-likelihood ratio for hypothesis testing in paper \cite{MS-DC}. The agreement between Bayesian and Frequentist methods is not a coincident, because we have used uniform priors in our derivation. 

Now we are ready to use the result in \cite{MS-DC} for estimating the expected difference between the log-ICLs,
\begin{align}
\label{eq:normalization}
&\mathbb{E}\left[\ln P_{DC}(G ,\hat{g}\mid M_i)-\ln P_{SBM}(G ,\hat{g}'\mid M_i')\right]\nonumber\\
=& \mathbb{E}[\Lambda_{DC}(G,g)] +\Theta(\frac{n}{2}\ln \frac{n}{k})-\Theta(n\ln \frac{2|E|}{n})\nonumber\\
\approx& \ln [(\frac{1}{2}+\frac{n}{24|E|})(n-k)]+\Theta(\frac{n}{2}\ln \frac{n}{k})-\Theta(n\ln \frac{2|E|}{n})\;,
\end{align}
where $\bar{d}_{s}$ is the average degree of vertices in block $s$ and $\frac{n}{24|E|}$ is the first order correction for sparse graphs \cite{MS-DC}. 

Equation \eqref{eq:normalization} only holds when the vanilla SBM is the generative model of the data. For datasets in general, however, we can guarantee this happen by making sure the number of blocks $k$ is so large that even the vanilla SBM over-fits. Going back to Figure \ref{fig:dcDistVdc}, the red curves peaks at $k=10$. We can then normalized the log-likelihood of the DC-SBM by subtracting it with the expected difference (Equation \eqref{eq:normalization}) at $k=10$.

Subtracting Equation \eqref{eq:normalization} from \eqref{eq:BIC_DC_derive}, we have the normalized posterior of the DC-SBM which is now directly comparable to Equation \eqref{eq:sumBayes}/\eqref{eq:sumBayesSparse}:
\begin{align}
\label{eq:posterior_norm}
&\ln P_{DC}(G ,g\mid M_i) = \ln P(G,g\mid\hat{\theta},\hat{q},\hat{p})\nonumber\\
			&- \frac{k^2}{2}\ln \Theta(n^*) - \ln [(\frac{1}{2}+\frac{n}{24|E|})(n-k)]\;.
\end{align}

Normalization allows us to compare multiple models with different $k$ together. Now we can ask questions like which model should we use conditioned on a given number of blocks. If we compare the two models for the same $k$ according to Figure \ref{fig:dcDistVdc}, the choice would be DC-SBM when $k<3$, and the vanilla SBM when $k>6$ (There is no clear winner when $3\le k\le 6$). The best model overall is a vanilla SBM with 10 blocks, which is consistent with our generative model (5 blocks with bimodal degree distributions).

To arrive at the BIC formulation, we discard the constants in Equation \eqref{eq:posterior_norm} and multiply both sides by $-2$,
\begin{align}
\label{eq:BIC_DC}
\textit{BIC}_{DC}(M_i) =& - 2\ln P(G,g|M_i, \hat{\Pi}_i)\nonumber\\
			&+ k^2\ln \Theta(n^*) + 2\ln[\Theta(n)]\;.
\end{align}

Compared with the vanilla SBM (Equation \eqref{eq:BIC_dense}/\eqref{eq:BIC_sparse}), $\textit{BIC}_{DC}$ has an additional penalty term which grows with the size of the network. Therefore, DC-SBM favors fewer blocks than the vanilla SBM, since the flexibility provided by the additional parameters allow vertices with very different degrees to coexist in the same block. 

\subsection{Results on real world networks}
To illustrate how the universal framework might work on real world data sets, we investigate two simple social networks using Equation \eqref{eq:BIC_dense} and \eqref{eq:posterior_norm}. The first is a social network consisting of $34$ members of a karate club, where undirected edges represent friendships \cite{zachary}. The network is made up of two assortative blocks, each with one high-degree hub and many low-degree peripheral vertices. In \cite{MS-DC}, the authors studied the model selection problem between the vanilla SBM and DC-SBM conditioned on $k=2$. Using the frequentist likelihood ratio test, there were not enough evidence to reject the null hypothesis that the network is generated by a vanilla SBM. 

\begin{figure}
\begin{center}
\includegraphics[width=.4\textwidth]{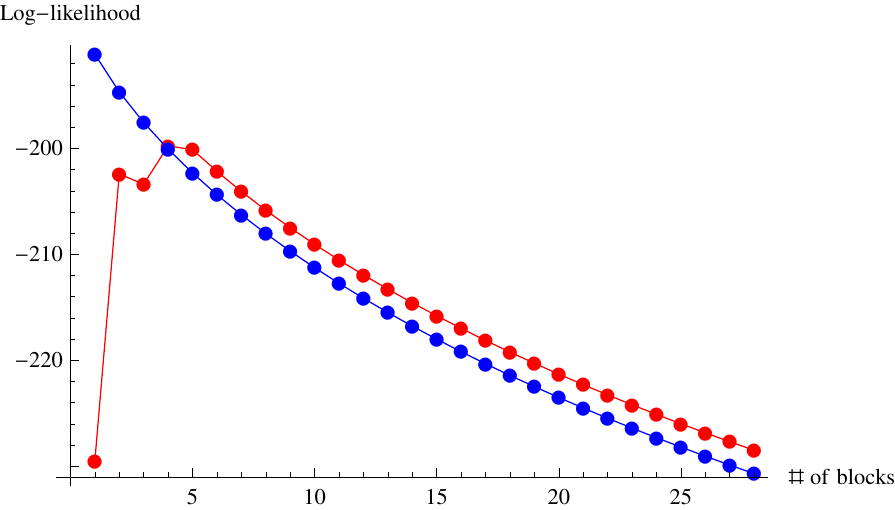}
\includegraphics[width=.4\textwidth]{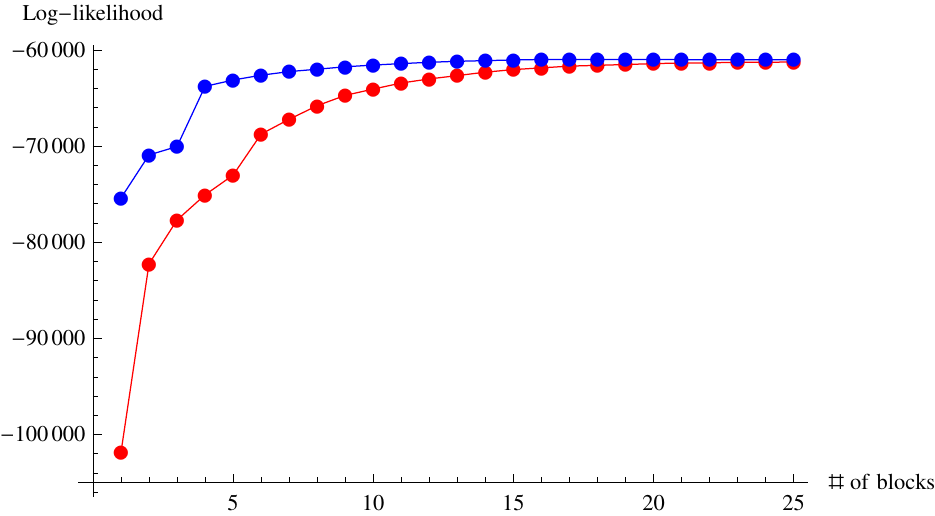}
\end{center}
\caption{The change of log-ICL for the karate club network (top) and the political blogs network (bottom) as the number of blocks grows. Both the vanilla SBM (red) and DC-SBM (blue) with different $k$ are fitted to the data. The experiment is done using a Monte Carlo sampling method. The log-likelihood values shown here has been normalized for both models.}
\label{fig:realNets}
\end{figure}

The result based on the log-ICL with normalization confirms that the DC-SBM has a higher likelihood at $k=2$ (see Figure~\ref{fig:realNets}, top). However, for DC-SBM, it maximizes at $k=1$, which means with degree correction, any blocking leads to over-fitting. Therefore, for any meaningful communities, the better choice is the vanilla SBM with a bigger $k$, because the degree heterogeneity is not strong enough to justify the DC-SBM. In fact, the best model according to the universal Bayesian framework is the vanilla SBM with $k=4$, which corresponds to the division with high/low degree blocks for each cluster. This is also reminiscent to the result using active learning \cite{moore_active}, where the vanilla SBM labels most of the vertices correctly once the high degree vertices are known. 

The second example is a network of political blogs in the US \cite{adamic_pBlogs}. Here we focus on the largest component with 1222 blogs and 19087 links between them. The blogs have known political leanings, with either liberal or conservative labels. The network is assortative, with heavy tailed degree distributions within each block. As a consequence, the frequentist analysis in \cite{MS-DC} supported the hypothesis that the network is generated by a DC-SBM. The universal Bayesian framework confirms the previous conjectures based on frequentist arguments at $k=2$ (see Figure \ref{fig:realNets}, bottom). In fact, the degree heterogeneity here is so strong, that the vanilla SBM never overtake DC-SBM even with very big $k$ values. If you are interested in large scale community structure of the political blogs network, such as the political factions, it seems the DC-SBM with smaller $k$ values is a more reasonable choice.

\section*{Acknowledgment}
The authors would like to thank Cristopher Moore and Tiago Peixoto for valuable inputs and discussions.



%
\bibliographystyle{IEEEtran}
\bibliography{Reference}

\end{document}